\documentclass{article} %
\usepackage{arxiv}

\usepackage[utf8]{inputenc} %
\usepackage[T1]{fontenc}    %
\usepackage{hyperref}       %
\usepackage{url}            %
\usepackage{booktabs}       %
\usepackage{amsfonts}       %
\usepackage{nicefrac}       %
\usepackage{microtype}      %
\usepackage{amsmath}
\usepackage{multirow}
\usepackage{algorithm,algpseudocode,float}

\title{Training Sparse Neural Networks using\\ Compressed Sensing}

\author{
  Jonathan W. Siegel \\
  Department of Mathematics\\
  Pennsylvania State University\\
  University Park, PA 16802 \\
  \texttt{jus1949@psu.edu} \\
  \And 
  Jianhong Chen \\
  Department of Mathematics\\
  Pennsylvania State University\\
  University Park, PA 16802 \\
  \texttt{jxc5788@psu.edu} \\
  \And 
  Pengchuan Zhang \\
  Microsoft Research\\
  \texttt{penzhan@microsoft.com}
  \AND
  Jinchao Xu \\
  Department of Mathematics\\
  Pennsylvania State University\\
  University Park, PA 16802 \\
  \texttt{jxx1@psu.edu} \\
}

\begin{document}

\maketitle

\begin{abstract}
 Pruning the weights of neural networks is an effective and widely-used technique for reducing model size and inference complexity. We develop and test a novel method based on compressed sensing which combines the pruning and training into a single step.  Specifically, we utilize an adaptively weighted $\ell^1$ penalty on the weights during training, which we combine with a generalization of the regularized dual averaging (RDA) algorithm in order to train sparse neural networks. The adaptive weighting we introduce corresponds to a novel regularizer based on the logarithm of the absolute value of the weights. We perform a series of ablation studies demonstrating the improvement provided by the adaptive weighting and generalized RDA algorithm. Furthermore, numerical experiments on the CIFAR-10, CIFAR-100, and ImageNet datasets demonstrate that our method 1) trains sparser, more accurate networks than existing state-of-the-art methods; 2) can be used to train sparse networks from scratch, i.e. from a random initialization, as opposed to initializing with a well-trained base model; 3) acts as an effective regularizer, improving generalization accuracy.
\end{abstract}

\section{Introduction}
Recently, neural networks have led to a breakthrough in a wide range of fields, especially computer vision and natural language processing \cite{lecun2015deep}. However, neural network models require large amounts of computational power both for training and inference. In addition, trained models often contain an extremely large number of parameters, which require a large amount of memory to store. This makes it difficult or even impossible to employ trained networks in computationally limited environments. Approaches to overcome this problem include quantizing the weights of neural networks \cite{courbariaux2015binaryconnect,hubara2017quantized,wu2018training,han2015deep,han2016eie,yin2019blended}, designing compressed architectures which are computationally less expensive yet achieve comparable accuracy \cite{howard2017mobilenets,tan2019efficientnet,iandola2016squeezenet}, and network pruning, which involves removing weights from a network as part of the training process \cite{han2015deep,han2016eie,han2015learning}.

We focus on the problem of network pruning, for which a very popular and effective approach is the iterative pruning and retraining method \cite{han2015learning}. This method works by pruning small weights after training, then finetuning or retraining the resulting model and repeating this process. There are also methods which combine the pruning and training into a single step \cite{ding2019global,alvarez2017compression,alvarez2016learning,dong2017learning}. While existing methods achieve impressive results, compressing networks by over an order of magnitude with minimal loss in accuracy, we demonstrate that by utilizing a weighted $\ell^1$-norm penalty motivated by compressed sensing \cite{donoho2006compressed} significantly smaller networks which achieve the same or better accuracy can be obtained.

Compressed sensing is concerned with the recovery of a sparse signal from a limited number of measurements. Importantly, the support of the signal is not known and must be inferred from the measurements. It is well-known that an $\ell^1$-norm regularizer can be used to recover the signal under a variety of assumptions \cite{tibshirani1996regression,candes2006robust,donoho2006compressed}. This suggests using an $\ell^1$-norm regularizer on the weights of a neural network to train sparse models. Indeed, the $\ell^1$-norm has been used successfully in neural network training, although an $\ell^2$-norm penalty combined with pruning and retraining often performs better for obtaining sparse networks\cite{han2015learning}. In addition, when seeking structured sparsity, the closely related group lasso regularizer \cite{yuan2006model} is often used effectively \cite{wen2016learning}.

In the field of compressed sensing, other regularizers have been shown to reconstruct sparser signals than the $\ell^1$-norm. For instance, an adaptively weighted $\ell^1$-norm \cite{candes2008enhancing} and the $\ell^p$-norm with $0\leq p < 1$\cite{chartrand2008iteratively}. These are also related, since the $\ell^p$-norm can be implemented via an iteratively reweighted $\ell^1$-norm penalty, although an iteratively reweighted $\ell^2$-norm approach is computationally simpler \cite{chartrand2008iteratively}. However, such techniques have not, to the best of our knowledge, been applied to the training of deep neural networks yet. In this work, we seek to fill this gap.

In particular, motivated by compressed sensing, we design an algorithm for training sparse neural networks which is based on an adaptively weighted $\ell^1$-norm penalty. There are two major problems we overcome in doing this. First, we design an appropriate adaptive $\ell^1$ weighting scheme for neural network training. Second, we design an algorithm which generates sparse iterates during training and also achieves good generalization accuracy. For convex objectives, the regularized dual averaging (RDA) algorithm \cite{xiao2010dual} was designed for this purpose, and we generalize it to the setting of deep learning. These contributions are summarized as follows.

\begin{itemize}
 \item We design an adaptively weighted $\ell^1$-regularization scheme which works well for training sparse neural networks. We connect this regularization scheme with a novel logarithmic regularizer, and also show how to adapt it to obtain structured sparsity.
 \item In order to use this scheme to train sparse neural networks, we propose the use of a modified version of the regularized dual averaging (RDA) method which incorporates momentum.
 \item Experimental results indicate that, on a variety of datasets and architectures, our method trains networks which generalize better and are significantly sparser than existing state-of-the-art methods.
\end{itemize}

\section{Related Work}
It is widely accepted that highly overparameterized neural networks are easier to train and achieve better generalization accuracy than smaller models \cite{bengio2006convex,hinton2015distilling,zhang2016understanding}. However, large models are also computationally expensive to deploy. This motivates the idea of network pruning, where most of the weights and complexity of the network are removed during training. This idea was first introduced in \cite{lecun1990optimal,hassibi1993second} and shown to be highly effective in \cite{han2015learning}. Since then, a variety of approaches to training sparse neural networks have been proposed.

\subsection{Unstructured pruning}
Many successful approaches to network pruning follow the pruning and retraining methodology first successfully used in \cite{han2015learning,han2015deep}. In this approach, weights with small magnitude are removed after training and the remaining weights are finetuned. This process is often repeated multiple times to compress networks with no loss in generalization accuracy. A variety of heuristics for pruning have been introduced, for instance based on neuron outputs instead of weight magnitudes \cite{hu2016network}, redundancy \cite{mariet2015diversity,srinivas2015data}, and second derivatives \cite{dong2017learning}. The difference between fine-tuning the remaining unpruned weights, retraining them from scratch, and retraining them from their initial values is studied in \cite{frankle2018lottery,liu2018rethinking,zhou2019deconstructing}, where specifically the ``Lottery Ticket Hypothesis'' is explored. Also, in \cite{han2015deep}, it is shown that pruning can be effectively combined with weight quantization to further improve network compression.

In addition, many methods which combine pruning and training have been proposed. In particular, it has been shown that variational dropout \cite{molchanov2017variational} can be used to train exceptionally sparse networks. An $\ell^1$-norm penalty on the weights combined with the RDA algorithm \cite{xiao2010dual} is applied to neural networks in \cite{he2018make}. In \cite{ding2019global} a modified momentum method is used to train sparse networks, which is particularly effective even for deep architectures. The use of momentum to obtain sparsity is also considered in \cite{dettmers2019sparse}. This paper builds upon these approaches by using an adaptively weighted $\ell^1$-norm regularizer to combine the pruning and training steps. An important difference between our method and most previous methods which combine training and pruning is that our algorithm trains the network from scratch, i.e. from a random initialization, as opposed to initializing with a well-trained base model. 

\section{Applying compressed sensing to neural network training}
We begin by introducing some notation. The neural networks which we consider consist of a sequence of convolutional, linear, and batch normalization \cite{ioffe2015batch} layers. Each of the convolutional and  linear layers contain trainable weights $W$ and bias parameters $b$. In addition, the batch normalization layers contain trainable shift and scale parameters $\gamma$ and $\beta$. We group the weights of the network by layer and also separate weights $W$ from biases $b$, and shifts $\beta$ from scale parameters $\gamma$ in convolutional, linear and batch normalization layers. We denote the resulting groups of parameters $G_1,...,G_N$, where each group $G_i$ is either $W$ or $b$ from a convolutional or linear layer, or is $\gamma$ or $\beta$ from a batch normalization layer. Here $N$ is the total number of groups, which will be twice the number of layers. We also let $\Theta = \{G_i\}_{i=1}^N$ denote the collection of all parameters, $\mathcal{D}$ denote the training dataset, and
\begin{equation}
 L(\Theta) = \frac{1}{|\mathcal{D}|}\sum_{(x,y)\in \mathcal{D}} l(x,y,\Theta)
\end{equation}
denote the empirical loss function, where $l$ is typically taken to be the cross-entropy. Finally, we denote by $\tilde{\nabla}L(\Theta)$ a stochastic gradient sample obtained by considering a mini-batch of training data.

\subsection{Adaptively weighting $\ell^1$-norm}
The lasso \cite{tibshirani1996regression}, which involves adding an $\ell^1$-norm regularization to the regression loss function, is a well-known and effective method for performing sparse regression and signal estimation in compressed sensing. In the context of neural network training, this corresponds to solving
\begin{equation}
 \arg\min_{\Theta} L(\Theta) + \lambda\|\Theta\|_1,
\end{equation}
where $\lambda$ is a hyperparameter controlling the trade-off between sparsity and training loss.

While the lasso regularizer can be effectively applied to neural network training, it can be considerably improved by using an adaptive $\ell^1$ weight. An adaptively reweighted $\ell^1$-constraint was first introduced in \cite{candes2008enhancing}, where it is shown to be considerably better at recovering sparse signals than the Danzig selector \cite{candes2007dantzig}. In this method, the $\ell^1$-norm on a parameter $\theta\in \Theta$ is weighted by $|\tilde\theta|^{-1}$, where $\tilde\theta$ is an estimate of the true value of $\theta$ obtained in the previous iteration. 

We use a similar adaptively weighted $\ell^1$-norm on the weights of our neural networks. Our method differs in the observation that parameters belonging to the same group $G_i$ should be of a similar scale, while parameters belonging to different $G_i$ can have very different scales. This suggests using a scheme similar to that introduced in \cite{candes2008enhancing} within each group $G_i$. Specifically, we weight the $\ell^1$-norm on a parameter $\theta\in G_i$ with
\begin{equation}\label{adaptive_weight}
 \lambda(\beta + 1)\left(\beta + \frac{|\theta|}{M_i}\right)^{-1},
\end{equation}
where $\beta$ and $\lambda$ are hyperparameters and $M_i$ is the maximum absolute value of all parameters in $G_i$, i.e.
$M_i = \max_{\theta\in G_i}|\theta|$.
This weight is $\lambda$ for the entry with largest absolute value in each group and increases to a maximum of $\lambda(1 + \beta^{-1})$ for entries which are $0$. So the hyperparameter $\lambda$ controls the total strength of the $\ell^1$ penalty, while $\beta$ controls how large the penalty can become for small weights. 

The precise form of the weighting scheme \ref{adaptive_weight} was determined through trial and error by running experiments on the LeNet-5 model on MNIST \cite{lecun1998gradient}. However, it can also be related to a novel regularizer, via an argument given in \cite{chartrand2008iteratively}. There, an $\ell^p$-penalty for $0\leq p < 1$ is implemented using an adaptively weighted $\ell^1$-penalty, with appropriately weights. Following this viewpoint, the weighting scheme in \ref{adaptive_weight} can be understood as corresponding to the logarithmic regularizer
\begin{equation}\label{adaptive_weight_regularizer}
 R(\Theta) = \lambda(\beta + 1)\sum_{i=1}^N \sum_{\theta\in G_i}M_i\log\left(\beta + \frac{|\theta|}{M_i}\right),
\end{equation}
where $x = G_1\cup\cdots \cup G_n$ represents all of the weights in the network. This is very similar to the case of an $\ell^0$ penalty in \cite{chartrand2008iteratively}.

Finally, in order to make the algorithm more robust, we calculate the weights in \eqref{adaptive_weight} based upon a running average of the magnitudes of the parameters. In particular, we consider a running average of the  absolute values of each parameter, computed recursively according to
\begin{equation}\label{theta_averaging}
 |\theta|_n^{av} = \mu|\theta|_{n-1}^{av} + (1-\mu)|\theta_n|.
\end{equation}
Here $\mu$ is a momentum parameter which effectively controls the number of iterations over which we average, and we discuss it in more detail later. We use the averaged absolute values $|\theta|_n^{av}$ in \eqref{adaptive_weight} to calculate the adaptive weights in each iteration.

\subsection{Extended Regularized Dual Averaging}
A key difficulty in using an $\ell^1$-norm regularizer for neural network training is that the naive forward-backward stochastic gradient descent algorithm applied to the regularized loss function
\begin{equation}
 f(\Theta) = L(\Theta) + \lambda \|\Theta\|_1
\end{equation}
doesn't generate sparse iterates, which takes the form
\begin{equation}\label{forward_backward_sgd}
\begin{split}
    &\Theta_{n+\frac{1}{2}} = \Theta_{n} - s_n\tilde\nabla L(\Theta_n)\\
    &\Theta_{n+1} 
    \in \Theta_{n+\frac{1}{2}} - s_n\partial(\lambda \|\cdot\|_1)(\Theta_{n+1}),
\end{split}
\end{equation}
where $s_n$ is the step size in $n$-th iteration. Here the second line represents a backward step for the regularizer $\lambda\|\cdot\|_1$, whose solution can be given in closed form and is known as the soft-thresholding operator
 \begin{equation}\label{l1_prox}
 x_{n+1} = \arg\min_y \left(\lambda\|y\|_1 + \frac{1}{2}\|x_{n+\frac{1}{2}} - y\|_2^2\right) = \text{sign}(x_{n+\frac{1}{2}})\max(0,|x_{n+\frac{1}{2}}| - s_n\lambda).
\end{equation}
The fact that iteration \eqref{forward_backward_sgd} doesn't generate sparse iterates, which was first observed in \cite{xiao2010dual}, occurs because a small step size $s_n$ is often necessary for convergence, or to obtain good generalization accuracy, and this means that the soft-thresholding parameter $s_n\lambda$ is very small. This phenomenon has also been observed in the context of training compressed neural networks, where it has been termed magnitude plateau \cite{ding2019global}.

To overcome this issue for convex machine learning problems, the regularized dual averaging algorithm (RDA), which is an extension of the dual averaging algorithm of Nesterov \cite{nesterov2009primal}, is introduced in \cite{xiao2010dual}. A special case of this algorithm, which bears resemblance to \eqref{forward_backward_sgd}, is
\begin{equation}\label{rda_leap_frog}
\begin{split}
    &\Theta_{n+\frac{1}{2}} = \Theta_{n-\frac{1}{2}} - s_n\tilde\nabla L(\Theta_n)\\
    &\Theta_{n+1} 
    \in \Theta_{n+\frac{1}{2}} - S_n\partial(\lambda \|\cdot\|_1)(\Theta_{n+1}),
\end{split}
\end{equation}
where the backward step is given by $S_n = \sum_{k=1}^n s_i$ and $\Theta_{-\frac{1}{2}} = \Theta_0$. Note that this implies that the soft-thresholding parameter, $\lambda S_n$, grows significantly and this leads to sparse iterates. Extensive experiments showing the effectiveness of this method on convex problems are given in \cite{xiao2010dual}.

The RDA algorithm has previously been applied to the training of neural networks in \cite{he2018make}. However, when applied to highly non-convex problems such as neural network training, the RDA algorithm is not particularly robust and may not always converge. This is due to the fact that the soft-thresholding parameter increases too rapidly and this makes the algorithm highly sensitive to the initialization. To get around this issue, we use a generalization of the RDA algorithm, first proposed in \cite{siegel2019extended} and called xRDA (eXtended RDA). This algorithm introduces an additional averaging parameter $\alpha_n$ and takes the form
\begin{equation}\label{xrda_leap_frog}
\begin{split}
    &\Theta_{n+\frac{1}{2}} = (1-\alpha_n)\Theta_n + \alpha_n\Theta_{n-\frac{1}{2}} - s_n\tilde\nabla L(\Theta_n)\\
    &\Theta_{n+1} 
    \in \Theta_{n+\frac{1}{2}} - S_n\partial(\lambda \|\cdot\|_1)(\Theta_{n+1}),
\end{split}
\end{equation}
where the backward step size now satisfies $S_n = \alpha_n S_{n-1} + s_n$ and again $\Theta_{-\frac{1}{2}} = \Theta_0$. Notice that if $\alpha_n = 0$ we get SGD \eqref{forward_backward_sgd}, and if $\alpha_n = 1$ we get RDA \eqref{rda_leap_frog}. Thus this method can be thought of as interpolating between the two. In \cite{siegel2019extended} it is shown that this method converges for convex objectives. By slowly increasing $\alpha_n$ during the course of training, the soft-thresholding parameter slowly increases, improving the robustness of the algorithm and leading to better accuracy and sparsity.

\subsection{Momentum for RDA}
Finally, in order to improve the generalization accuracy of the trained sparse models, we introduce a method of adding momentum to RDA.
 Momentum, which has an intuitive physical interpretation \cite{polyak1963gradient}, has long been used to accelerate convex optimization \cite{nesterov1983method} and has been used extensively in deep learning to accelerate convergence and improve generalization accuracy \cite{sutskever2013importance}. In the context of deep learning, training with momentum typically means the search direction in each iteration is an exponentially weighted average of the gradient over the previous iterations, i.e.
\begin{equation}\label{classical_momentum}
 \begin{split}
    &v_n = \mu v_{n-1} + \tilde\nabla L(\Theta_n)\\
    &\Theta_{n+1} = \Theta_n - s_nv_n,
\end{split}
\end{equation}
where the momentum parameter $\mu$ effectively determines how many previous iterates the gradient is averaged over. We note that in the context of convex optimization, the above iteration must be modified to guarantee convergence for arbitrary convex objectives \cite{nesterov1983method}. A detailed investigation of \eqref{classical_momentum} and Nesterov's momentum for training deep neural networks is given in \cite{sutskever2013importance}.

We incorporate momentum in the xRDA algorithm \eqref{xrda_leap_frog} by replacing the sampled gradient $\tilde\nabla L(\Theta)$ by an average over the past gradients. This results in the algorithm
\begin{equation}\label{xrda_leap_frog_momentum}
\begin{split}
    &v_n = \mu v_{n-1} + (1-\mu)\tilde\nabla L(\Theta_n)\\
    &\Theta_{n+\frac{1}{2}} = (1-\alpha_n)\Theta_n + \alpha_n\Theta_{n-\frac{1}{2}} - s_n\tilde v_n\\
    &\Theta_{n+1} 
    \in \Theta_{n+\frac{1}{2}} - S_n\partial(\lambda \|\cdot\|_1)(\Theta_{n+1}),
\end{split}
\end{equation}
where as before the backward step size now satisfies $S_n = \alpha_n S_{n-1} + s_n$ and again $\Theta_{-\frac{1}{2}} = \Theta_0$.

The momentum parameter $\mu$ in \eqref{xrda_leap_frog_momentum} controls the scale over which we average our gradients to determine an appropriate search direction. This parameter is often taken independent of the step size, which corresponds to averaging the gradients over a fixed number of iterations. However, we have observed that better generalization accuracy is obtained if the gradients are averaged over a fixed timescale. In particular, we take the momentum parameter $\mu = e^{-s_n/T}$, where $T$ determines the time scale over which we average. We note that this dependence of momentum on step size corresponds to a discretization of Langevin dynamics with constant damping parameter \cite{bhattacharya2009stochastic}.

We also use this value of $\mu$ to determine the average parameter magnitudes $|\theta|_n^{av}$. In other words, we set $\mu = e^{-s_n/T}$ in equation \eqref{theta_averaging} so that the parameter magnitudes are averaged over a fixed time scale when calculating the adaptive $\ell^1$ weights. We have found that this significantly improves the stability and robustness of the algorithm.

\section{Experimental Results}

In this section, we provide experimental evidence demonstrating the effectiveness of our compressed sensing based approach to training sparse neural networks\footnote{Our code is available at \href{https://github.com/jwsiegel2510/xrda-sparse-training}{https://github.com/jwsiegel2510/xrda-sparse-training}}.

\subsection{Comparison with Existing State-of-the Art Methods}

We evaluate the effectiveness of using compressed sensing to train sparse neural networks on several common benchmark models and datasets. Specifically, we consider the datasets CIFAR-10 and CIFAR-100 \cite{krizhevsky2009learning}, and ImageNet \cite{deng2009imagenet}. In all of our tests, we run our algorithm from scratch, i.e. we do not need to initialize with a well-trained base model, unlike previous approaches which combine training and pruning into a single step \cite{ding2019global,molchanov2017variational}.
\subsubsection{CIFAR-10}
We present the results of our first set of experiments on CIFAR-10 in Table \ref{sparse-training-cifar10}. We test a variety of architectures, specifically VGG-16 and VGG-19 \cite{simonyan2014very}, and ResNet-18 and ResNet-56 \cite{he2016deep}. We note that for the ResNet-18 structure we use is much wider that the ResNet-56 network. In particular, the ResNet-56 architecture we use contains $3$ blocks with $[16,32,64]$ channels, while the ResNet-18 architecture contains $4$ blocks with $[64,128,256]$ channels. We use these two architectures to show that compressed sensing can be applied both to deep narrow and to shallow wide residual networks.

For VGG-16 and VGG-19 we set the momentum timescale $T = 9.5$ and use a cosine learning rate schedule \cite{loshchilov2016sgdr} with initial learning rate $1.0$. We have obtained the best results by setting the averaging parameter $\alpha_n = 0$, which corresponds to a version of stochastic gradient descent, although similar results can be obtained with a slowly increasing averaging parameter $\alpha_n$. Finally, we set the $\ell^1$-norm parameters $\lambda = 10^{-6}$ and $\beta = 2\cdot 10^{-3}$. For ResNet-16 and ResNet-18 we use mostly the same hyperparameters, but we set the initial learning rate to $0.25$ and set $\lambda = 1.3\cdot 10^{-6}$. We run the algorithm for $600$ epochs in all cases. This follows the methodology in \cite{ding2019global}, and we have found that this large number of epochs is necessary to give the algorithm time to find advantageous sparse configurations. The baseline accuracy is obtained by training with SGD using appropriately tuned hyperparameters for the same number of epochs. For all models, we use a standard data augmentation strategy, consisting of padding to $40$ by $40$, followed by a random crop back to $32$ by $32$, and random right-left flipping.

We can see from these results that using compressed sensing (CS) results in significantly sparser and more accurate networks than existing state-of-the-art methods. For instance, we are able to reduce the number of parameters in VGG-19 by more than two orders of magnitude while increasing the accuracy. In addition, this holds across a wide range of architectures, including both deep and wide residual neural networks.

\subsubsection{CIFAR-100}
Next, we present results on the CIFAR-100 dataset in Table \ref{sparse-training-cifar100}. Here we again consider the VGG-19 network architecture. We used the same hyperparameters as for VGG-19 on CIFAR-10 and the same data augmentation strategy, discussed previously.

We see from these results that compressed sensing (CS) is also effective on more complicated datasets. In particular, we are able to reduce the size of VGG-19 by a factor of nearly $40$ while attaining a significant increase in accuracy. This suggests that compressed sensing is an effective regularizer in addition to producing sparse networks.

\begin{table}[!htbp]
 \caption{Unstructured sparsity results on CIFAR-10.}
 \label{sparse-training-cifar10}
 \centering
 \begin{tabular}{l c c c c c c}
  \hline\multirow{2}{*}{Model} & \multirow{2}{*}{Algorithm} & Base & Sparse & Dense / Sparse  & Compression & Non-Zero\\ & & Top1 & Top1 & Parameters & Ratio & Fraction\\\hline
      \textbf{ResNet-18} & \textbf{CS}   & \textbf{95.05} &    \textbf{94.49} & \textbf{11.17M / 0.14M} &    \textbf{81x} &    \textbf{1.23}   \\
   ResNet-18 &  RDA \cite{he2018make}  &  - & 93.95& 11.17M / 0.56M &  20x    &  5.00   \\\hline
                \textbf{ResNet-20} & \textbf{CS}   & \textbf{93.87} &    \textbf{91.99} & \textbf{270K / 27K} &    \textbf{10x} &    \textbf{9.88}   \\
   ResNet-20 &  Bayesian \cite{deng2019adaptive}  &  93.90 & 91.68 & 270K / 27K &  10x    &  10.00   \\\hline
    \textbf{VGG-16} & \textbf{CS}   & \textbf{93.79} &    \textbf{94.13} & \textbf{14.73M / 0.18M} &    \textbf{80x} &    \textbf{1.25}   \\
  VGG-16 & Momentum \cite{dettmers2019sparse}   &  93.41 & 93.31& 14.73M / 0.74M &  20x    &  5.00   \\
   VGG-16 & Bayesian \cite{louizos2017bayesian}   &  91.60 & 91.00& 14.73M / 0.81M &  18x    &  5.50   \\
     VGG-16 & Var Dropout \cite{molchanov2017variational}   &  92.70 & 92.70& 14.73M / 0.31M &  48x    &  2.08   \\
   VGG-16 & Slimming \cite{liu2017learning}   &  93.66 & 93.41& 14.73M / 0.65M &  22x    &  4.40 \\ VGG-16 & DST \cite{junjie2019dynamic} & 93.74 & 93.36 & 14.73M / 1.4M & 10x & 10.00 
   \\ VGG-16 & DST \cite{junjie2019dynamic} & 93.74 & 93.00 & 14.73M / 0.74M & 20x & 5.00\\\hline
  \textbf{VGG-19} & \textbf{CS}   & \textbf{93.60} &    \textbf{94.18} & \textbf{20.04M / 0.19M} &    \textbf{104x} &    \textbf{0.97}   \\
  VGG-19 & Pruning \cite{han2015learning}   &  93.50 & 93.34& 20.04M / 1.00M &  20x    &  5.00   \\
   VGG-19 & Scratch-B \cite{liu2018rethinking}   &  93.50 & 93.63& 20.04M / 1.00M &  20x    &  5.00   \\\hline
 \end{tabular}
\end{table}

\begin{table}[!htbp]
 \caption{Unstructured sparsity results on CIFAR-100.}
 \label{sparse-training-cifar100}
 \centering
 \begin{tabular}{l c c c c c c}
  \hline\multirow{2}{*}{Model} & \multirow{2}{*}{Algorithm} & Base & Sparse & Dense / Sparse  & Compression & Non-Zero\\ & & Top1 & Top1 & Parameters & Ratio & Fraction\\\hline
  \textbf{VGG-19} & \textbf{CS}   & \textbf{73.83} &    \textbf{75.93} & \textbf{20.09M / 0.46M} &    \textbf{43x} &    \textbf{2.30}   \\
  VGG-19 & Pruning \cite{han2015learning}   &  71.70 & 70.22& 20.09M / 1.00M &  20x    &  5.00   \\
   VGG-19 & Scratch-B \cite{liu2018rethinking}   &  71.70 & 72.08& 20.09M / 1.00M &  20x    &  5.00   \\\hline
   \end{tabular}
\end{table}

\subsubsection{ImageNet}

Here we present the results of using compressed sensing to compress ResNet-50 \cite{he2016deep} on the ImageNet \cite{deng2009imagenet} dataset. The results can be found in table \ref{sparse-training-imagenet}. We see from this that compressed sensing significantly outperforms both pruning \cite{liu2018rethinking} and global sparse momentum \cite{ding2019global}. It is certainly competitive with and even slightly outperforms the state-of-the-art rigging the lottery method \cite{evci2019rigging}.

\begin{table}[!htbp]
\caption{Unstructured sparsity results on ImageNet.}
\label{sparse-training-imagenet}
\centering
\begin{tabular}{l c c c c c c}
 \hline\multirow{2}{*}{Model} & \multirow{2}{*}{Algorithm} & Base & Sparse & Dense / Sparse  & Compression & Non-Zero\\ & & Top1 & Top1 & Parameters & Ratio & Fraction\\\hline
 \textbf{ResNet-50} & \textbf{CS}   & \textbf{} &    \textbf{76.98} & \textbf{25M / 3.5M} &    \textbf{7.14x} &    \textbf{14.00}   \\
 \textbf{ResNet-50} & \textbf{CS}   & \textbf{} &    \textbf{76.00} & \textbf{25M / 1.7M} &    \textbf{14.7x} &    \textbf{6.8}   \\
 ResNet-50 & Pruning \cite{liu2018rethinking}   &  76.15 & 76.06 & 25M / 17.5M &  1.43x    &  70.00\\   
 ResNet-50 & Pruning \cite{liu2018rethinking}   &  76.15 & 76.09 & 25M / 10M &  2.5x    &  40.00\\
 ResNet-50 & RigL \cite{evci2019rigging}   &  76.8 & 77.1 & 25M / 5M &  5x    &  20.00\\
 ResNet-50 & RigL \cite{evci2019rigging}   &  76.8 & 76.4 & 25M / 2.5M &  10x    &  10.00\\
 ResNet-50 & GSM \cite{ding2019global}   &  75.72 & 75.33 & 25M / 6.25M &  4x    &  25.00\\
 ResNet-50 & GSM \cite{ding2019global}   &  75.72 & 74.30 & 25M / 5M &  5x    &  20.00\\
 \\\hline
  \end{tabular}
\end{table}

\section{Conclusion and future work}
We have shown that compressed sensing can be used to effectively train sparse neural networks. In particular, by developing an appropriate adaptively weighted $\ell^1$-norm penalty, combined with a novel optimization algorithm based on the RDA algorithm, we are able to train significantly smaller and sparser networks than previously possible on both CIFAR-10 and CIFAR-100. A significant increase in generalization accuracy also suggests that compressed sensing has an important regularizing effect.

Further work involves more extensive testing and tuning of the algorithm, in addition to a theoretical analysis of compressed neural networks. In particular, we propose testing this technique on larger and more complex datasets such as ImageNet \cite{deng2009imagenet}, and attempting to obtain better results with more extensive hyperparameter tuning. We also propose testing different types of structured sparsity using the method, and expect that this technique will also be effective for different types of machine learning models beyond feed-forward convolutional neural networks. Finally, we hope that a theoretical analysis can explain the precise nature of the regularizing effect of compressed sensing observed in our experiments.

\section{Acknowledgements}
We would like to thank Dr. Xiao Lin for helpful comments and discussion while preparing this work.

\bibliographystyle{spmpsci}\bibliography{refs} 

\end{document}